\documentclass[journal]{IEEEtai}

\usepackage[colorlinks,urlcolor=blue,linkcolor=blue,citecolor=blue]{hyperref}
\usepackage{color,array,amsmath,bm}
\usepackage[ruled, linesnumbered, vlined]{algorithm2e}
\usepackage{amsfonts} 

\usepackage{graphicx}
\usepackage{overpic}
\usepackage{soul} 
\usepackage{graphicx}
\usepackage{multirow}
\usepackage{booktabs}
\usepackage[a4paper, total={7in, 10in}]{geometry}
\usepackage[font=footnotesize]{caption}
\makeatletter
\newcommand{\removelatexerror}{\let\@latex@error\@gobble}
\makeatother

\begin{document}

\title{Brain-inspired Evolutionary Architectures for Spiking Neural Networks} 
\author{Wenxuan Pan, Feifei Zhao,  Zhuoya Zhao, Yi Zeng

\thanks{Wenxuan Pan is with the Brain-inspired Cognitive Intelligence Lab, Institute of Automation, Chinese Academy of Sciences, Beijing 100190, China, and School of Artificial Intelligence, University of Chinese Academy of Sciences, Beijing 100049, China.} 
\thanks{Feifei Zhao is with the Brain-inspired Cognitive Intelligence Lab, Institute of Automation, Chinese Academy of Sciences, Beijing 100190, China.} 
\thanks{Zhuoya Zhao is with the Brain-inspired Cognitive Intelligence Lab, Institute of Automation, Chinese Academy of Sciences, Beijing 100190, China, and School of Future Technology, University of Chinese Academy of Sciences, Beijing 101408, China.} 

\thanks{Yi Zeng is with the Brain-inspired Cognitive Intelligence Lab, Institute of Automation, Chinese Academy of Sciences, Beijing 100190, China, and State Key Laboratory of Multimodal Artificial Intelligence Systems, Institute of Automation, Chinese Academy of Sciences, Beijing 100190, China, and University of Chinese Academy of Sciences, Beijing 100049, China, and Center for Excellence in Brain Science and Intelligence Technology, Chinese Academy of Sciences, Shanghai 200031, China.}

\thanks{The first and the second authors contributed equally to this work, and serve as co-first authors.}
\thanks{The corresponding author is Yi Zeng (e-mail: yi.zeng@ia.ac.cn).}}

\maketitle

\begin{abstract}
The complex and unique neural network topology of the human brain formed through natural evolution enables it to perform multiple cognitive functions simultaneously.
Automated evolutionary mechanisms of biological network structure inspire us to explore efficient architectural optimization for Spiking Neural Networks (SNNs).
Instead of manually designed fixed architectures or hierarchical Network Architecture Search (NAS), this paper evolves SNNs architecture by incorporating brain-inspired local modular structure and global cross-module connectivity. Locally, the brain region-inspired module consists of multiple neural motifs with excitatory and inhibitory connections; Globally, we evolve free connections among modules, including long-term cross-module feedforward and feedback connections. 
We further introduce an efficient multi-objective evolutionary algorithm based on a few-shot performance predictor, endowing SNNs with high performance, efficiency and low energy consumption. 
Extensive experiments on static datasets (CIFAR10, CIFAR100) and neuromorphic datasets (CIFAR10-DVS, DVS128-Gesture) demonstrate that our proposed model boosts energy efficiency, archiving consistent and remarkable performance.
This work explores brain-inspired neural architectures suitable for SNNs and also provides preliminary insights into the evolutionary mechanisms of biological neural networks in the human brain.
\end{abstract}


\section{Introduction}

After millions of years of evolution, neurons with complex information processing capabilities and microcircuits with specific functions have emerged in the human brain, empowering it to be a powerful device.
As the third generation of neural network, Spiking Neural Network (SNN)~\cite{maass1997networks} realizes a low energy consumption and high-efficiency computing paradigm by simulating the characteristics of neuron communication in the brain.
However, existing SNN research employs fixed architectures that lack references to brain-inspired topological properties, limiting their performance on multiple tasks.

Neural Architecture Search (NAS) attempts to replace the experience-based manual design of architectures with an automated searching approach. Studies have shown that it even outperforms architectures designed with the expertise of human experts on some tasks~\cite{real2019regularized,zoph2018learning}.
As a kind of optimization scheme in NAS, Evolutionary Neural Architecture Search (ENAS) is promising due to its insensitivity to local optimal values and without a large number of computing resources compared to reinforcement learning (RL)-based algorithms~\cite{liu2021survey,zhang2021adaptive,xie2022benchenas}. Based on the current understanding of the evolutionary structure of the brain, this paper is committed to exploring the evolution of brain-inspired local and global neural circuits for optimizing SNN architecture and function.

In ENAS, when the population and the number of iterations are large, it is very time-consuming to train SNN models one by one to evaluate the performance of individuals. Thus efficient evaluation methods are particularly important, and research ideas are usually inseparable from weight-sharing~\cite{bender2018understanding,guo2020single}, performance predictor~\cite{sun2019surrogate,wen2020neural} or zero-shot methods~\cite{abdelfattah2021zero,mellor2021neural}. The weight-sharing method is also called the one-shot method, and only one supernet needs to be trained in the entire algorithm process. However, training to convergence takes a long time, and it may encounter the problem of performance collapse, which could not reflect real network fitness. Zero-cost evaluation methods require no training and have extremely low time costs, but only roughly measuring attributes positively related to architectural performance may also lead to inaccurate rankings of fitness. As a few-shot method, performance predictor trains a regression model for directly predicting the performance of unknown architectures based on the information of some trained architectures~\cite{sun2019surrogate,wen2020neural,xie2023efficient, deng2017peephole,liu2018progressive} and can be divided into online and offline according to whether the predictor is updated during the NAS process. Offline predictors only sample limited architectures once, and no new information is considered after construction. Therefore, the online predictor is more practical and flexible, able to continuously improve the evaluation accuracy by leveraging the existing knowledge~\cite{liu2018progressive, wei2022npenas, lu2020nsganetv2}.

A large amount of ENAS research is based on deep neural networks (DNNs), and existing works on evolving SNNs architecture are still very limited.
\cite{na2022autosnn} proposes a search model named AutoSNN to evolve more energy-efficient architecture. It defines an artificially formulated weight factor to optimize the number of spikes and classification accuracy, which is still essentially a single-objective optimization without manifesting the coordination of the two contradictory evolutionary objectives. 
~\cite{kim2022neural} applies a zero-shot approach to evaluate the classification performance of SNNs, greatly reducing the time and computational cost. 
In addition, from the perspective of search space, \cite{na2022autosnn} follows the topological characteristics of VGG-Net and ResNet, searching for the stacking method of spiking convolution and residual blocks. \cite{kim2022neural} considers the cell-based search space with only one global backward connection.
The setting of such search space following the experience of DNN does not make full use of the brain-inspired structure or module and also limits the search efficiency of the evolutionary algorithm.
SNNs originated from the exploration of the computing nature of the brain, which inspired us to delve deeper into evolutionary SNN architecture from a more biologically interpretable perspective.

After millions of years of evolution, the synaptic connection patterns between nearly 100 billion different excitatory and inhibitory neurons in the human brain have formed diverse circuit motifs and modular brain regions~\cite{luo2021architectures,meredith2002neuronal}.
Discoveries in neuroscience encourage us to design a neural architecture search space more suitable for SNN from a brain-inspired perspective: locally, each module incorporates 5 common types of circuit motifs found in different brain regions; globally, non-hierarchical long-term connections including cross-module or feedback connections are evolved between modules, resembling the complex connections between different brain regions. At the same time, in order to speed up the evolutionary process, we employ a few-shot multi-objective evolutionary algorithm, which can continuously promote the evolution of the population towards the direction of low energy consumption and high efficiency without sampling and training a large number of candidates.

Overall, the proposed Evolutionary Brain-inspired Neural Architecture Search (EB-NAS) aims to accelerate the process of multi-objective evolutionary algorithms through an online performance predictor, and automatically design low-energy, high-efficiency SNN models combined with brain-inspired topological properties. 
All contributions can be divided into the following points:
\begin{itemize}

\item[1)] We evolve the SNNs architectures based on brain-inspired region modules consisting of neural motifs, and the free long-term connections (such as cross-module, feedback) between these modules. Such a search space dictates that SNNs could automatically evolve into a more biologically plausible architecture comparable to the brain.

\item[2)] 
To search for high-performance and low-energy SNNs architectures, we use a multi-objective evolutionary algorithm (using as few spikes as possible to achieve higher performance). We employ a time-efficient few-shot evaluation method, an online performance predictor, to alleviate the time-consuming evolution process.

\item[3)] 
On a variety of datasets, our method achieves state-of-the-art (SOTA) performance with significantly lower energy consumption. 
The thorough analysis also proves the effectiveness of drawing on the naturally evolved neural topology of the human brain for evolving SNNs architectures.

\end{itemize}

The remainder of the paper is organized as follows.  Section \uppercase\expandafter{\romannumeral2} introduces some related works about ENAS. Section \uppercase\expandafter{\romannumeral3} elaborates the implementation details of our proposed algorithm. Section \uppercase\expandafter{\romannumeral4} verifies the superiority of the proposed model from various perspectives, and Section \uppercase\expandafter{\romannumeral5} summarizes the main contributions of this paper and gives an outlook.

\section{Related Work}

NAS is a method for automatically designing neural network architectures, which has been proven in many studies~\cite{elsken2019neural, ghiasi2019fpn,zoph2016neural} to be more efficient than manually designed architectures based on human expert experience. Existing NAS algorithms can be roughly divided into three categories: RL-based methods, gradient-based methods and evolution-based methods. RL-based methods consume a lot of computing resources~\cite{liu2021survey,ren2021comprehensive} and require an additional controller to help search. 
Gradient-based methods need to build a supernet in advance, making it not fully automatic and often likely to discover ill-conditioned architectures. 

ENAS is a computing paradigm that simulates species evolution or population behaviour. The candidate architectures will be initialized and encoded, then optimized and iterated through some evolutionary computing (EC) methods such as genetic algorithms (GAs)~\cite{mitchell1998introduction}, particle swarm optimization (PSO)~\cite{kennedy1995particle}, evolutionary strategy (ES)~\cite{back1997handbook}, and differential evolution (DE)~\cite{price2006differential} until the fitness of the population is gradually evolved to obtain a set of high-quality solutions. ENAS is more computationally resource-efficient than RL-based NAS and requires less prior knowledge than gradient-based NAS, can be widely used to solve complex multi-objective optimization problems~\cite{lu2020nsganetv2,sun2018igd,lu2019nsga}.  

\subsection{Multi-objective ENAS}

Many ENAS algorithms simultaneously pursue performance and cost~\cite{lu2019nsga,lu2020multiobjective,elsken2018efficient,yang2020cars}, and these conflicting objective functions bring challenges to iterative optimization. Common optimization methods for dealing with multi-objectives can be divided into scalarization and population-based methods~\cite{lu2020nsganetv2}. The scalarization methods~\cite{na2022autosnn,cai2018proxylessnas,howard2019searching,tan2019mnasnet,wu2019fbnet} integrate multiple objectives into an objective function with a fixed weight factor. ~\cite{na2022autosnn} considers both the number of spikes (evaluating SNN energy consumption) and classification accuracy with scalarized fitness function. The problem with such methods is that the artificially pre-defined preference of different goals before the search is not accurate enough, it does not take into account the coordination between two potentially contradictory evolution goals. 
Population-based approaches~\cite{elsken2018efficient,chu2021fairnas,lu2020muxconv} use some heuristic search strategy instead of random sampling. They can more fully consider each objective than scalarization methods, thus approaching the Pareto optimal front for multiple objectives.

\subsection{Efficient Evaluation}
During evolution, the performance of the current architecture needs to be evaluated to indicate the direction of next-generation optimization. But training a large neural network from scratch until convergence requires many GPU days~\cite{real2019regularized,zoph2016neural,real2017large}. Thus evaluation is often the most time-consuming stage during the evolution process. How to shorten the evaluation time has become a concern of many studies.
Based on the number of architectures that need to be trained, efficient evaluation methods can be divided into three categories: few-shot, one-shot, and zero-shot methods.

One-shot and zero-shot require only one or zero training of the architecture to obtain the performance of all architectures.
However, the one-shot methods~\cite{chu2021fairnas,zhang2022evolutionary} train a supernet that includes all candidate architectures. This training process takes a lot of time to converge even if it does not collapse, and may cause multi-model forgetting problems~\cite{benyahia2019overcoming}. Even with reduced computation time compared to training each candidate from scratch, the accuracy of the supernet's predictive power is not necessarily reliable. 
Without training any architectures, zero-shot methods propose many architecture-related indicators from the perspective of neural network theory, such as learnability~\cite{mellor2021neural}, generalization~\cite{xu2021knas} and expressiveness~\cite{lin2021zen}, but these rough measurements also lead to inaccurate reflection of candidate performance~\cite{xie2023efficient}.

Performance predictor~\cite{sun2019surrogate,wen2020neural,xie2023efficient, deng2017peephole} is a few-shot method, which obtains the architecture and its corresponding accuracy by sampling and training a small number of candidates and builds a performance regression model that can predict unknown individuals.
A predictor that is automatically updated during construction and training is called an online predictor.~\cite{liu2018progressive} trains the predictor starting with the simplest shallow models, progressing to complex structures and culling bad structures. Promising architectures are used as new samples for the predictor, whose state is continuously updated.~\cite{lu2020nsganetv2} takes a similar approach to train adaptive switching surrogate predictors.
Compared with offline predictors, these works can continuously improve prediction performance with new samples (architecture and corresponding performance). 
\subsection{Search Space}
In addition to improving the speed of evaluation, researchers also optimized the search space for efficiency.
Block is proposed as a new topology, where the layers inside it have specific connections mode and functions, such as ResBlock~\cite{he2016deep}, DenseBlock~\cite{huang2017densely}, ConvBlock~\cite{elsken2017simple} and InceptionBlock~\cite{szegedy2015going}. The cell-based search space can be regarded as a special case of a block where all blocks are the same, but the connections of the layers within the block can be free~\cite{zoph2018learning,ying2019bench,dong2020bench}.
Existing works of searching the SNN architecture have not made more innovations in the search space.~\cite{na2022autosnn} draws on the experience of ConvBlock and ResBlock to find the best combination of them.~\cite{kim2022neural} follows the cell-based search space, but only focuses on one backward connection, making the search capability limited. All of them lack deep reference to the brain topology properties, and the performances need to be improved.

In view of the above-mentioned deficiency of NAS on SNN, this paper optimizes the search space and evaluation method from a brain-inspired perspective.
A more efficient and biologically interpretable search space is developed, in which modules consisting of five common circuit motifs and the long-term global connectivity between them can be evolved.
Besides, a few-shot online performance predictor is built to speed up the population-based multi-objective evolution process without loss of accuracy.
\section{Methods}

\subsection{Spiking Neural Network Foundation}

\begin{figure}[htp]
\centering
\includegraphics[width=8cm]{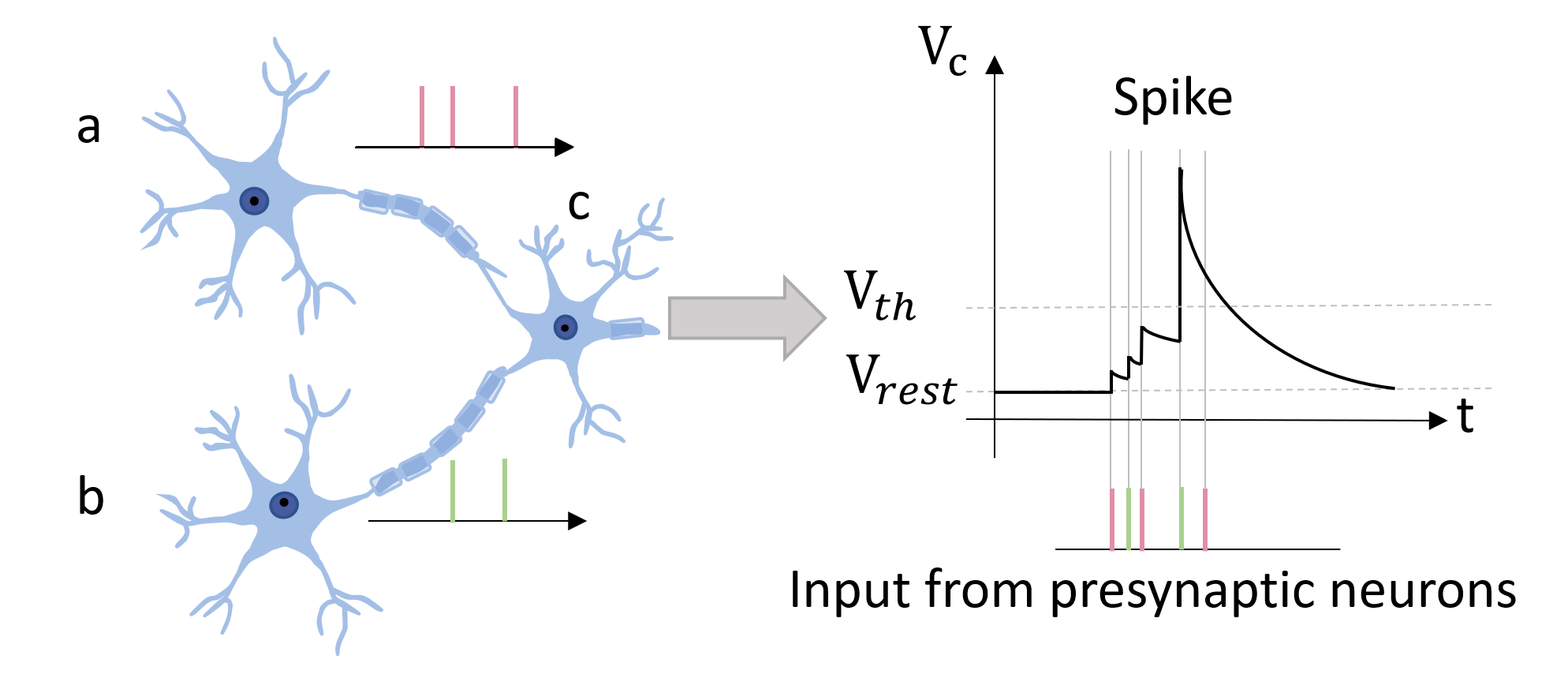}
\caption{Computation of LIF neurons. Neuron $c$ receives spikes from presynaptic neurons $a$ and $b$, and the potential $V_c$ accumulates continuously from the resting potential $V_{rest}$ until it reaches the threshold $V_{th}$, fires a spike and returns to the resting potential.}
\label{lif}
\end{figure}

In SNNs, information-carrying spikes are transmitted between pre- and post-synaptic neurons. The post-synaptic neuron fires when the firing of the pre-synaptic neuron causes the potential of the post-synaptic neuron to accumulate until it reaches a threshold. Here, we adapt a commonly used Leaky Integrate-and-Fire (LIF) neuron model~\cite{lapicque1907recherches}. The dynamic process of neuron $i$ membrane potential $V_i(t)$ changes with time $t$ can be described as Eq.~\eqref{eq1} and Fig.~\ref{lif}.

\begin{figure*}[htp]
\centering
\includegraphics[width=18cm]{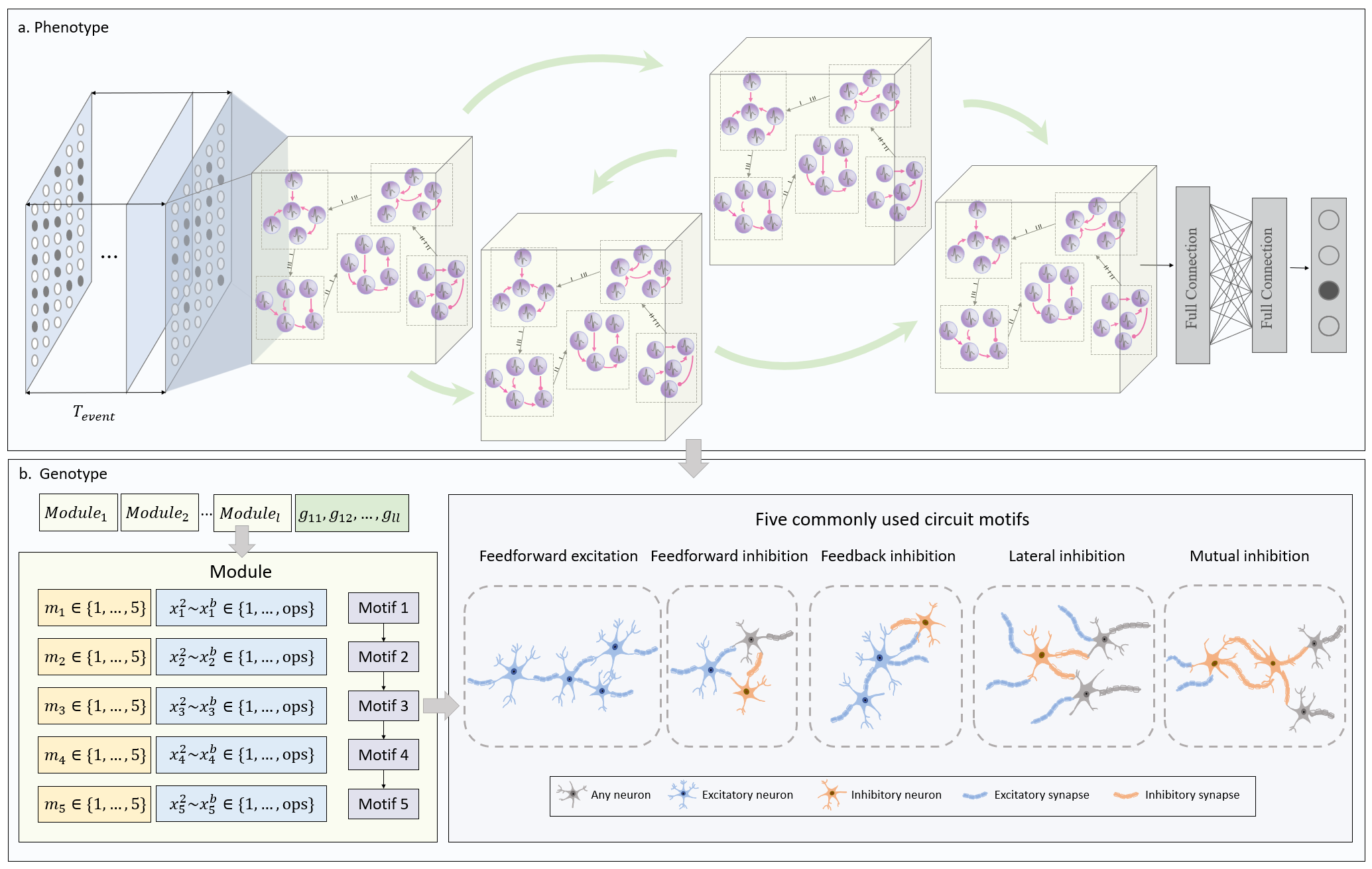}
\caption{Phenotype-to-genotype mapping and encoding schemes. a. The entire network is composed of $l$ modules, each of which contains 5 neural circuits. Globally, the modules are freely connected without self-loop. b. The encoding scheme for the model.}
\label{encode}
\end{figure*}
\begin{align}
\label{eq1}
\tau_m \frac{\mathrm{d} V_i(t)}{\mathrm{d} t}=-\left[V_i(t)-V_{rest }\right]+I_i(t)
\end{align}%

\begin{align}
\label{eq2}
h(V_i(t))=\left\{\begin{array}{l}
1, V_i(t) \geq V_{th} \\
0, V_i(t)<V_{th}
\end{array}\right.
\end{align}
where $\tau_m=RC$ is the membrane potential time constant ($R$ is the membrane resistance, $C$ is the membrane capacitance). 
The input current $I_i(t)$ leads to the accumulation of membrane potential $V_i(t)$. In Eq.~\eqref{eq2}, $h(V_i(t))$ is a function indicating the firing of neuron $i$.
When the threshold $V_{th}$ is reached, the neuron fires and the potential returns to $V_{rest}$. This calculation process is repeated for a certain period of time, and the number of spikes accumulated by the last layer of neurons during this period will directly affect the output of the model.

The spikes transfer process described in Eq.~\eqref{eq2} is non-differentiable, which limits the optimization of network weights by backpropagation. Thus we train SNNs with a surrogate gradient~\cite{wu2018spatio} as Eq.~\eqref{eq3}:

\begin{align}
\label{eq3}
\frac{\partial h(V_i(t))}{\partial V_i(t)}= \begin{cases}1 & \text { if } -\frac{1}{\alpha} \leq V_i(t)-V_{t h} \leq \frac{1}{\alpha} \\ 0 & \text { otherwise. }\end{cases}
\end{align}
where $\alpha$ (set to 2) is a parameter that determines the steepness of the curve. The spiking neuron model and training algorithm used in this work are implemented based on Brain-inspired Cognitive Intelligence Engine (BrainCog) framework~\cite{zeng2023braincog}.

\subsection{Encoding Brian-inspired Neural Circuit}
\label{encodedblock}

In our model, the search space for evolving SNN architectures includes local modular structure consisting of five neural motifs, and global free long-term connectivity between modules. 
Common neural circuit motifs found in the human brain are Feedforward Excitation (FE), Feedforward Inhibition (FI), Feedback Inhibition (FbI), Lateral Inhibition (LI) and Mutual Inhibition (MI). For example, FE and FbI are related to multisensory fusion~\cite{meredith2002neuronal} and enhancing the signal-to-noise ratio~\cite{luo2021architectures}. FI helps the network maintain a balance of excitatory and inhibitory~\cite{isaacson2011inhibition}. LI plays a role in amplifying the perception of various sensory stimuli~\cite{laurent2001odor} and MI is linked to the rhythmic activity of neural networks~\cite{grillner2006biological,saper2010sleep}. 
We implement them with spiking neurons as shown in Fig.~\ref{encode}b. 
As shown in Fig.~\ref{encode}a, 
the motif combinations in different modules are various, which is inspired by the different structures of the brain regions.
Global connections imitate the way brain regions are connected, not a conventional hierarchical structure but freely connected, without restricting the direction of the connection and whether it crosses modules.

\begin{align}
\label{genomesnn}
{[{Module}_1,{Module}_2,{Module}_3...,{Module}_l,g_{11},g_{12},..,g_{ll}]}
\end{align}%

A single module $Module_i$ consisting of five motifs is encoded as:
\begin{align}
\label{block}
{[m_1,x_1^2,...,x_1^{b},m_2,x_2^2,...,x_2^{b},...,m_5,x_5^2,...,x_5^{b}]}
\end{align}%
where each motif is coded as $(m_i,x_i^2,...,x_i^{b})$. $m_i$ indicates which kind of circuit motif it is, while $x_i^2$ to $x_i^{b}$ encode the computation inside the motif. $x_i^2$ to $x_i^{b}$ have $ops$ possible values, and each represents one kind of operation. $b$ is the maximum number of genes computed internally within a motif and set to 20.
Here, we experimentally selected two convolution operations: 3x3 convolution and 5x5 convolution.

$[g_{11},g_{12},..,g_{ll}] $ are global parameters indicating the connection mode between $l$ modules. For example, $g_{ij}$ indicates whether there is a connection from $Module_i$ to $Module_j$. In order to prevent self-loop, we limit $g_{ii}$ to 0. See Table~\ref{gene} for a detailed description of the meaning of the different gene values in a genome.

\begin{table}[htp]
\begin{tabular}{lll}
    \toprule

Gene & Value & Description \\
    \midrule
 \multirow{2}{*}{$g_{ij}$} & $0$ & No connection from $Module_i$ to $Module_j$. \\

& $1$ & One connection from $Module_i$ to $Module_j$. \\
    \midrule
 & 1 & Feedforward Excitation \\
& 2 & Feedforward Inhibition \\
$m_i$ & 3 & Feedback Inhibition \\
& 4 & Lateral Inhibition \\
& 5 & Mutual Inhibition \\

    \midrule
 \multirow{2}{*}{$x_i^2 \sim x_i^b$} & 1 & $3 \times 3$ conv \\
& 2 & $5 \times 5$ conv \\

\bottomrule

\end{tabular}
\caption{Detailed description of the meanings of different gene values in a genome.}
\label{gene}
\end{table}

\begin{figure}[t]
\centering
\includegraphics[width=8.5cm]{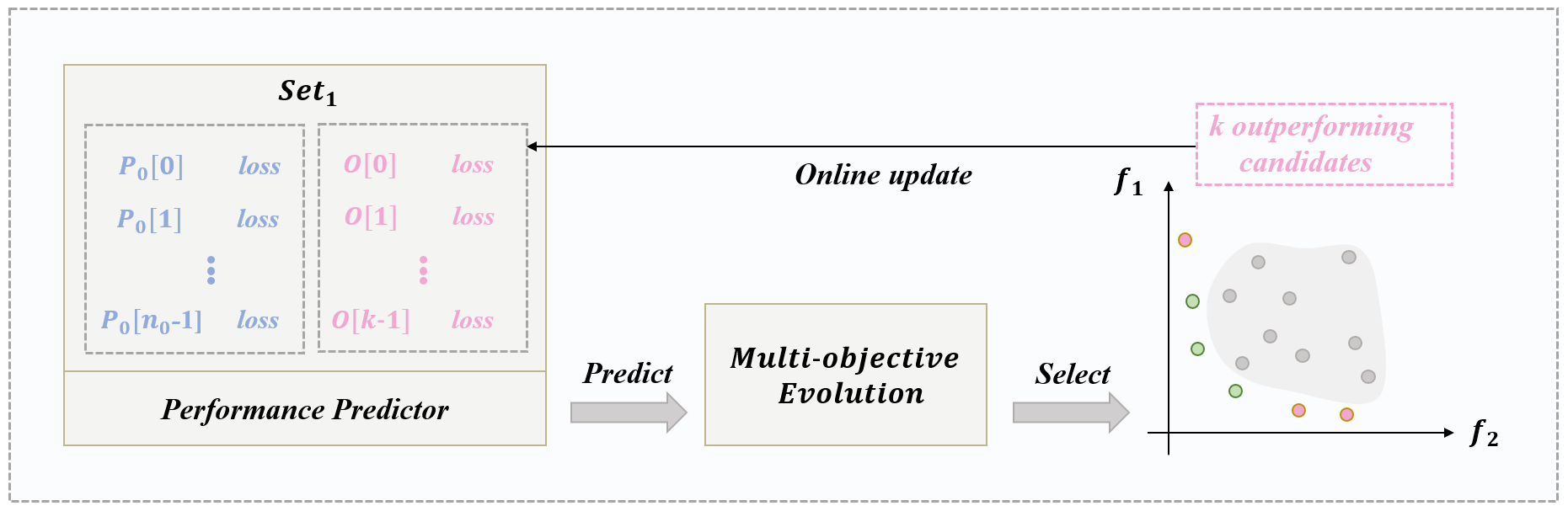}
\caption{The multi-objective evolutionary process with the help of the performance predictor. Blue represents the initial population and pink indicates the selected $k$ outperforming candidates after non-dominated sorted, which will be trained to obtain real loss and used to further enrich the knowledge of the predictor.}
\label{predictor}
\end{figure}

\subsection{Evolutionary Neural Architecture Search for SNN}

\subsubsection{Multi-objective Optimization}

For SNNs, the improvement of accuracy is often accompanied by the loss of energy, that is, more spikes are fired. In order to find an SNN architecture with lower energy consumption and higher efficiency, this paper takes both the validation error and the number of spikes as optimization goals. This multi-objective optimization problem (MOP) can be expressed as:

\begin{align}
\label{mop}
\underset{M \in \Omega}{\operatorname{argmin}} \boldsymbol{F}(M) =\left\{ f_1(M), f_2(M)\right\}
\end{align}
The SNN architecture $M$ must be within the specified search space $\Omega$ consisting of all genomes encoded based on Eq.~\ref{genomesnn} and Table~\ref{gene}.
The first objective is to minimize the verification error of the model $M$ on a given image classification dataset:
\begin{equation}
f_1(M)=\min \mathrm{L}\left(M, \mathcal{D}_{\text {tst }} , \mathcal{D}_{\text {trn }}\right)
\end{equation}
where $\mathrm{L}(*)$ represents the verification error of the model $M$ trained on $\mathcal{D}_{\text {trn }}$ on the validation dataset $\mathcal{D}_{\text {tst }}$.

Suppose the model $M$ consists of $l$ modules, then the second evolutionary goal can be expressed as:
\begin{equation}
f_2(M)=\sum_{i=0}^l S(Module_i)
\end{equation}
where $\mathrm{S}(Module_i)$ indicates the number of spikes fired by all spiking neurons in $Module_i$.


\begin{algorithm}
    \SetKwData{Left}{left}
    \SetKwData{This}{this}
    \SetKwData{Up}{up} 
    \SetKwFunction{Union}{Union}
    \SetKwFunction{FindCompress}{FindCompress} 
    \SetKwInOut{Input}{input}
    \SetKwInOut{Output}{output} 
    
    \Input{initial population size $n_0$,  number of iterations to search $iters$, number of generations to evolve in NSGA-\uppercase\expandafter{\romannumeral2} $gens$, training dataset $\mathcal{D}_{\text {trn }}$, validation dataset $\mathcal{D}_{\text {tst }}$,
      number of epochs required for evaluation $e_{eval}$,
    number of epochs required for training $e_{trn}$, search space $\Omega$.} 
    \Output{Accuracy $acc$ of the evolved candidate.} 
        
        {$P_0$ $\gets$ Initialize a population of size $n_0$ in $\Omega$}\;
        {$Set_1$ $\gets$ $\emptyset$}\;

      \While {$i<n_0$}{
        {$loss$ $\gets$ $f_1$($P_0[i]$, $\mathcal{D}_{\text {trn }}$,$\mathcal{D}_{\text {tst }}$, $epochs$=$e_{eval}$)}\;
        {$Set_1$ $\cup$ $(P_0[i],loss)$ }\;

    }

      \While {$t<iters$}{
        {$Predictor$ $\gets$ UpdatePredictor($Set_1$)}\;

        {$O$ $\gets$ NSGA-\uppercase\expandafter{\romannumeral2}($Predictor$,$n_{new}$,$gens$,$f_2$,$k$)}\;

    \For{$j\leftarrow 0$ \KwTo $k$}{
        {$loss$ $\gets$ $f_1$($O[j]$, $\mathcal{D}_{\text {trn }}$,$\mathcal{D}_{\text {tst }}$, $epochs$=$e_{eval}$)}\;
        {$Set_1$ $\cup$ $(O[j],loss)$ }\;
        }
    }
    {Select the non-dominated individual with the lowest loss in $O$ and train $e_{trn}$ epoch to obtain the top-1 accuracy $acc$}\;
\Return{$acc$}
\caption{The procedure of EB-NAS algorithm.}
\label{alg:evo} 
\end{algorithm}

\subsubsection{Efficient Evaluation}
We want to find non-dominated architectures of the entire search space $\Omega$,
however, there are too many architectures to evaluate during the evolution process, which is also the most time-consuming part~\cite{liu2021survey}.
Therefore, we build an online performance predictor $Predictor$ based on $Set_1$: $Predictor$ is a Classification And Regression Trees (CART)~\cite{sun2019surrogate} surrogate model that can quickly predict the validation error of unknown architectures based on the existing knowledge set $Set_1$, avoiding training all searched architectures and improving evolutionary efficiency.

At initialization, an initial population $P_0$ of size $n_0$ is constructed. Each individual in $P_0$ can be trained for $e_{eval}$ epochs to obtain a corresponding verification error through $f_1$, which is stored in $Set_1$. 
In order to make the search of the evolutionary algorithm more directional, we take the following approach: in $t_{th}$ iteration, 
a new population of size $n_{new}$ in $\Omega$ is sampled, which can be evaluated by $Predictor$.
Use NSGA-\uppercase\expandafter{\romannumeral2} algorithm~\cite{deb2000fast} to implement two-point crossover and polynomial mutation operators, input $Predictor$ and $f_2$ to get the fitness of all individuals.
After $gens$ generations, the non-dominated sorted population is obtained. Each of $k$ outstanding candidates (denoted as $O$) will be trained for $e_{eval}$ epochs, and the validation error is added to $Set_1$, enriching the knowledge for training $Predictor$.
\begin{table}[htp]
\centering
\begin{tabular}{lll}
    \toprule

Parameter & Description &Value \\
    \midrule
 \multirow{10}{*}{}$\tau$ & membrane potential time constant & 2.0 \\
$V_{rest}$ & resting potential of neurons& 0 \\
$V_{th}$ & firing threshold of neurons& 0.5 \\
$l$ & maximum number of modules& 4 \\
 $n_0$ & initial population size& 300 \\
  $iters$ & number of iterations& 50 \\
$n_{new}$ & size of newly generated population in NSGA-\uppercase\expandafter{\romannumeral2}& 60 \\
 $gens$ & number of generations evolved in NSGA-\uppercase\expandafter{\romannumeral2} & 40\\
$e_{eval}$ & number of epochs required for evaluation& 10 \\
 $e_{trn}$ & number of epochs required for training& 600 \\
 $k$ & number of candidates selected in each iteration& 10 \\

\bottomrule

\end{tabular}
\caption{The value of all parameters in our model.}
\label{param}
\end{table}

\begin{table*}[htbp]
  \centering
  \caption{Comparison of classification performance on static datasets CIFAR10 and CIFAR100.  
  }
  \resizebox{6 in}{!}{
    \begin{tabular}{clcccc}
    \toprule
    \textbf{Dataset} & \textbf{Model} & \textbf{Training Methods} & \textbf{Architecture} & \textbf{Timesteps} & \textbf{Accuracy (\%)} \\
    \midrule
    \multirow{14}[4]{*}{CIFAR10} 
      & Wu et al.~\cite{wu2018spatio} & STBP & CIFARNet &12 & 89.83 \\
      & Wu et al.~\cite{wu2019direct} & STBP NeuNorm & CIFARNet &12 & 90.53 \\
      & Kundu et al.~\cite{kundu2021spike} &Hybrid &VGG16 &100 &91.29\\
      & Zhang \& Li~\cite{zhang2020temporal} & TSSL-BP & CIFARNet &5 & 91.41 \\
      & Shen et al.~\cite{shen2022backpropagation} & STBP & 7-layer-CNN &8 & 92.15 \\
      & Rathi et al.~\cite{rathi2020enabling} & Hybrid & ResNet-20 &250 & 92.22 \\
      & Rathi \& Roy~\cite{rathi2020diet} & Diet-SNN & ResNet-20 &10 & 92.54 \\
      & Li et al.~\cite{li2021free} &ANN-SNN Conversion &VGG16 &32 &93.00\\
      & Zheng et al.~\cite{zheng2021going} & STBP-tdBN & ResNet-19 &6 & 93.16 \\
      &Fang et al.~\cite{fang2021incorporating}&STBP &6Conv, 2Linear &8 &93.50\\
      & Han et al.~\cite{han2020rmp}&ANN-SNN Conversion &VGG16 &2048 &93.63\\
      & Deng et al.~\cite{deng2022temporal} & TET & ResNet-19 &4 & 94.44 \\
\cmidrule{2-6}      & Kim et al.~\cite{kim2022neural} & STBP & NAS &5 & 92.73 \\
      & Na et al.~\cite{na2022autosnn} & STBP & NAS &16 & 93.15 \\
      & \textbf{Our Method} & STBP & EB-NAS &\textbf{4} & \textbf{95.66} \\
    \midrule
    \multirow{12}[4]{*}{CIFAR100} 
    &Lu and Sengupta~\cite{lu2020exploring}&ANN-SNN Conversion &VGG15 &62 &63.20\\
    & Rathi \& Roy~\cite{rathi2020diet} & Diet-SNN & ResNet-20 &5 & 64.07 \\
    &Kundu et al.~\cite{kundu2021spike}&Hybrid &VGG11 &120 &64.98\\
    & Rathi et al.~\cite{rathi2020enabling} & Hybrid & VGG-11 &125 & 67.87 \\
    &Garg et al.~\cite{garg2021dct} &DCT &VGG9 &48 &68.30\\
    &Deng et al.~\cite{deng2021optimal}&ANN-SNN Conversion &ResNet-20  &32 &68.40\\
    & Park et al.~\cite{park2020t2fsnn}& TTFS &VGG15 &680 &68.80\\
    & Shen et al.~\cite{shen2022backpropagation} & STBP & ResNet34 &8 & 69.32 \\
    &Han et al. .~\cite{han2020rmp}&ANN-SNN Conversion &VGG16 &2048 &70.90\\
    &Li et al.~\cite{li2021free}  &ANN-SNN Conversion &ResNet-20  &16 &72.33\\
\cmidrule{2-6}   & Na et al.~\cite{na2022autosnn} & STBP & NAS &16 & 69.16 \\
        & Kim et al.~\cite{kim2022neural} & STBP & NAS &5 & 73.04 \\
      & \textbf{Our Method} & STBP & EB-NAS &\textbf{4} & \textbf{79.21} \\
    \bottomrule
    \end{tabular}}
  \label{stat}%
\end{table*}%
\begin{table*}[htbp]
  \centering
  \caption{Comparison of classification performance on neuromorphic datasets CIFAR10-DVS and DVS128-Gesture. 
}
    \resizebox{6 in}{!}{\begin{tabular}{clcccc}
    \toprule
    \textbf{Dataset} & \textbf{Model} & \textbf{Training Methods} & \textbf{Architecture} & \textbf{Timesteps} & \textbf{Accuracy (\%)} \\
    \midrule
    \multirow{7}[4]{*}{CIFAR10-DVS}  & Kugele et al.~\cite{kugele2020efficient}& Streaming Rollout & DenseNet &10 & 66.8 \\
    & Zheng et al.~\cite{zheng2021going} & STBP-tdBN & ResNet-19 &10 & 67.8 \\
     & Wu et al.~\cite{wu2021liaf} & BPTT & LIAF-Net &10 & 70.40 \\
      & Shen et al.~\cite{shen2022backpropagation} & STBP & 5-layer-CNN &16 & 78.95 \\
      & Deng et al.~\cite{deng2022temporal} & TET & VGGSNN &10 & 83.17 \\
\cmidrule{2-6}      & Na et al.~\cite{na2022autosnn} & STBP & NAS &16 & 72.50 \\
      & \textbf{Our Method} & STBP & EB-NAS &\textbf{10} & \textbf{83.48} \\
    \midrule
    \multirow{5}[3]{*}{DVS128-Gesture} & Xing et al~\cite{xing2020new} & SLAYER & 5-Layer-CNN &20 & 92.01 \\
      & Shrestha et al~\cite{shrestha2018slayer} & SLAYER & 16-layer-CNN &300 & 93.64 \\
      & Fang et al.~\cite{fang2021incorporating} & STBP & 5Conv, 2Linear &20 & 97.57 \\
      
\cmidrule{2-6}      & Na et al.~\cite{na2022autosnn} & STBP & NAS &16 & 96.53 \\
      & \textbf{Our Method} & STBP & EB-NAS &\textbf{10} & \textbf{98.08} \\
    \bottomrule
    \end{tabular}}
  \label{neuro}%
\end{table*}%

As the iterations proceed and more and more information in $Set_1$, $Predictor$ is trained in online mode to make it accurate.
Therefore, the found $O$ will get closer and closer to true non-dominated solutions in $\omega$.
The process demonstrated above is repeated until $t=iters$.
The non-dominated individual with the lowest loss in $O$ is selected and trained $e_{trn}$ epoch to get the top-1 accuracy $acc$ as the return value of the algorithm.
The whole process is shown in Algorithm~\ref{alg:evo} and Fig.~\ref{predictor}.
All parameter values involved in our model are shown in Table~\ref{param}.

\begin{figure*}[tp]
\includegraphics[width=18cm]{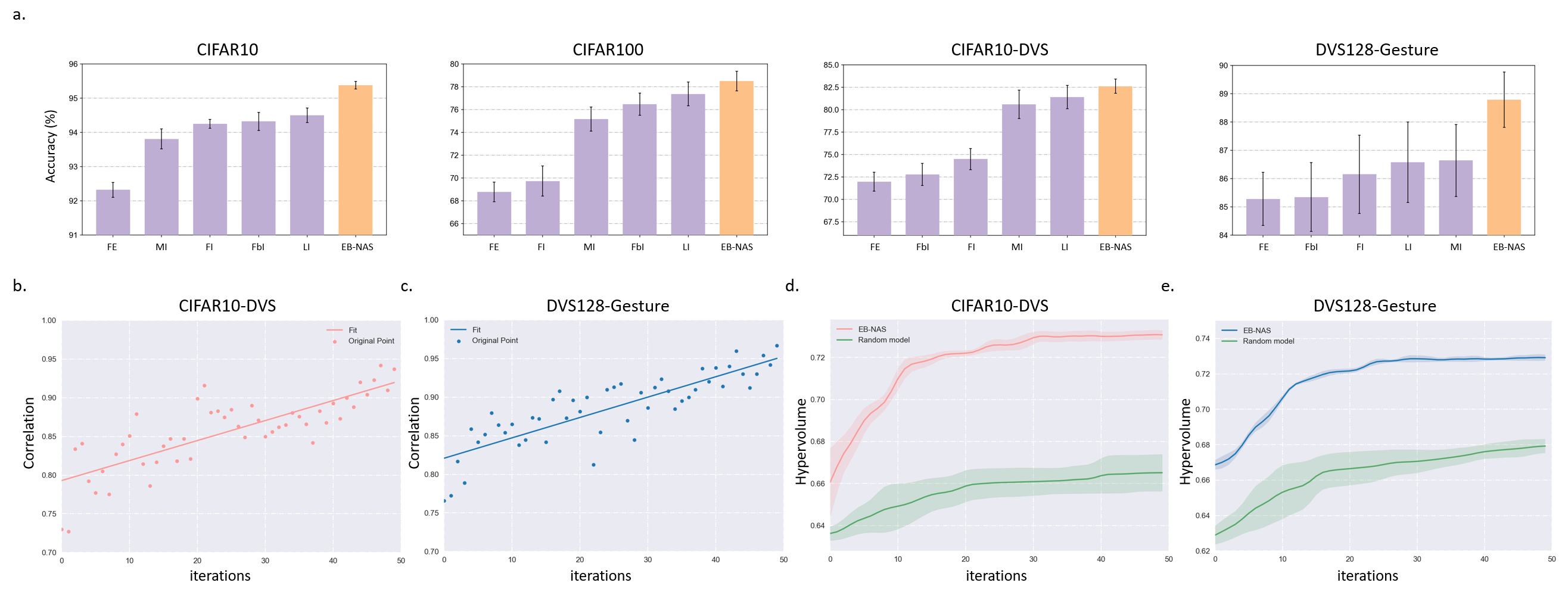}
\caption{Ablation results. a. Evolutionary performance comparison of single-motif models and EB-NAS. b. The accuracy of the predictor on CIFAR10-DVS in iteration. c. The accuracy of the predictor on DVS128-Gesture in iteration. d. The changes of hypervolume of two evolutionary models on CIFAR10-DVS. e. The changes of hypervolume of two evolutionary models on DVS128-Gesture.}
\label{abl}
\end{figure*}
\section{EXPERIMENTS AND ANALYSIS}

In this section, we conduct the experiments on static CIFAR10~\cite{krizhevsky2009learning}, CIFAR100~\cite{xu2015empirical} and neuromorphic DVS128-Gesture~\cite{amir2017low}, CIFAR10-DVS~\cite{li2017cifar10} datasets to illustrate the superiority of our proposed model. The neuromorphic dataset CIFAR10-DVS is an event stream-based dataset converted from the frame-based static dataset CIFAR10 by a Dynamic Vision Sensor (DVS).
DVS128-Gesture is a hand gesture dataset, which contains a combination of various lights and gestures and is recorded by a DVS for about 6 seconds. The classification performance and energy consumption of the evolved models (training for 600 epochs) are compared with multiple methods.

\subsection{Comparative Results}

Table~\ref{stat} shows the comparison results between our model and other SNN models on static datasets CIFAR10 and CIFAR100. 
On CIFAR10, compared with fixed-architecture SNN models, the classification accuracy of EB-NAS exceeds the best-performing ResNet-19 architecture~\cite{deng2022temporal} by 1.22\%. The performance of EB-NAS is also significantly better than models based on ResNet, VGG-Net and other convolutional structures, ranging from 2.03\% to 5.83\%.
Compared to existing SNN-based NAS models~\cite{na2022autosnn,kim2022neural} trained with STBP, the classification accuracy of EB-NAS is 2.51\% and 2.93\% higher, respectively.
On CIFAR100, EB-NAS outperforms the best fixed-architecture SNN model~\cite{li2021free} by at least 6.88\%. Compared with other SNN-based NAS models~\cite{na2022autosnn, kim2022neural}, our method improves the classification accuracy by 10.05\% and 6.17\% respectively. Among all the models listed in Table~\ref{stat}, EB-NAS uses the shortest simulation time (timesteps = 4), which greatly saves the parameter amount of the model.

Table~\ref{neuro} shows the comparison results between our model and other SNN models on neuromorphic datasets CIFAR10-DVS and DVS128-Gesture. 
On CIFAR10-DVS, the classification accuracy of our model exceeds that of fixed-structure models including DensNet and VGG-Net, ranging from 0.31\% to 16.68\%. Meanwhile, EB-NAS significantly outperforms the SNN-based NAS model~\cite{na2022autosnn} by 10.98\%. Similar conclusions can also be observed on DVS128-Gesture: whether it is a fixed structure or a structure searched by NAS, EB-NAS achieves better classification accuracy (with the shortest timesteps 10) than other SNN models.

In general, by comparing EB-NAS with fixed-structure SNNs and NAS-based SNNs on multiple datasets, we find that it is difficult for the existing NAS-based SNNs to exceed the performance of the fixed-designed architecture. EB-NAS surmounts this point, and its classification performance is not only higher than the existing NAS-based SNN models but also surpasses the manually designed SNN models. This shows that the classification performance of EB-NAS has reached the state-of-art and also demonstrates that SNN architectures automatically designed by NAS break through the optimization boundary of human expert experience, which is of great significance to the field of NAS for SNN.

\subsection{Ablation Study}
The brain-inspired search space customized for SNN proposed in this paper has two novel points: (1) using the combination of 5 functional microcircuits found in the brain as a basic module; (2) the global and free connectivity between modules, not limited to hierarchical connections. 
Ablation experiments will verify these two aspects.
Except for the variables mentioned in the ablation experiments, the parameters and process of the evolutionary algorithm are kept constant.
We conduct the ablation experiments on four datasets CIFAR10, CIFAR100, CIFAR10-DVS and DVS128-Gesture, and the initial channels are set to 48, 48, 32 and 32 respectively. 
\subsubsection{Combination of Motifs}

In order to verify the effectiveness of the combined motifs, we compare EB-NAS with models of fixed motif types and the results are shown in Fig.~\ref{abl}a. In each single-motif model, the motifs in all modules are specified as one of the five categories. In this way, five comparison models can be obtained, denoted as FE, FI, FbI, LI and MI in Fig.~\ref{abl}a. 
It can be seen from Fig.~\ref{abl}a that on CIFAR10, CIFAR100, CIFAR10-DVS and DVS128-Gesture, FE exhibits the worst classification ability, perhaps due to overexpression of excitation.
The performance of LI is relatively better, which may be due to the balance of excitatory and inhibitory properties inside.
Fig.~\ref{abl}a also reflects the superior classification accuracy of EB-NAS than that of other single-motif models. On all datasets, the classification accuracy of EB-NAS is 0.88\%, 1.13\%, 1.18\% and 2.15\% higher than the best single-motif model (average results of multiple trials), respectively, and shows the smallest variance on CIFAR10 and CIFAR10-DVS.

\begin{figure}[tp]
\centering
\includegraphics[width=7.5cm]{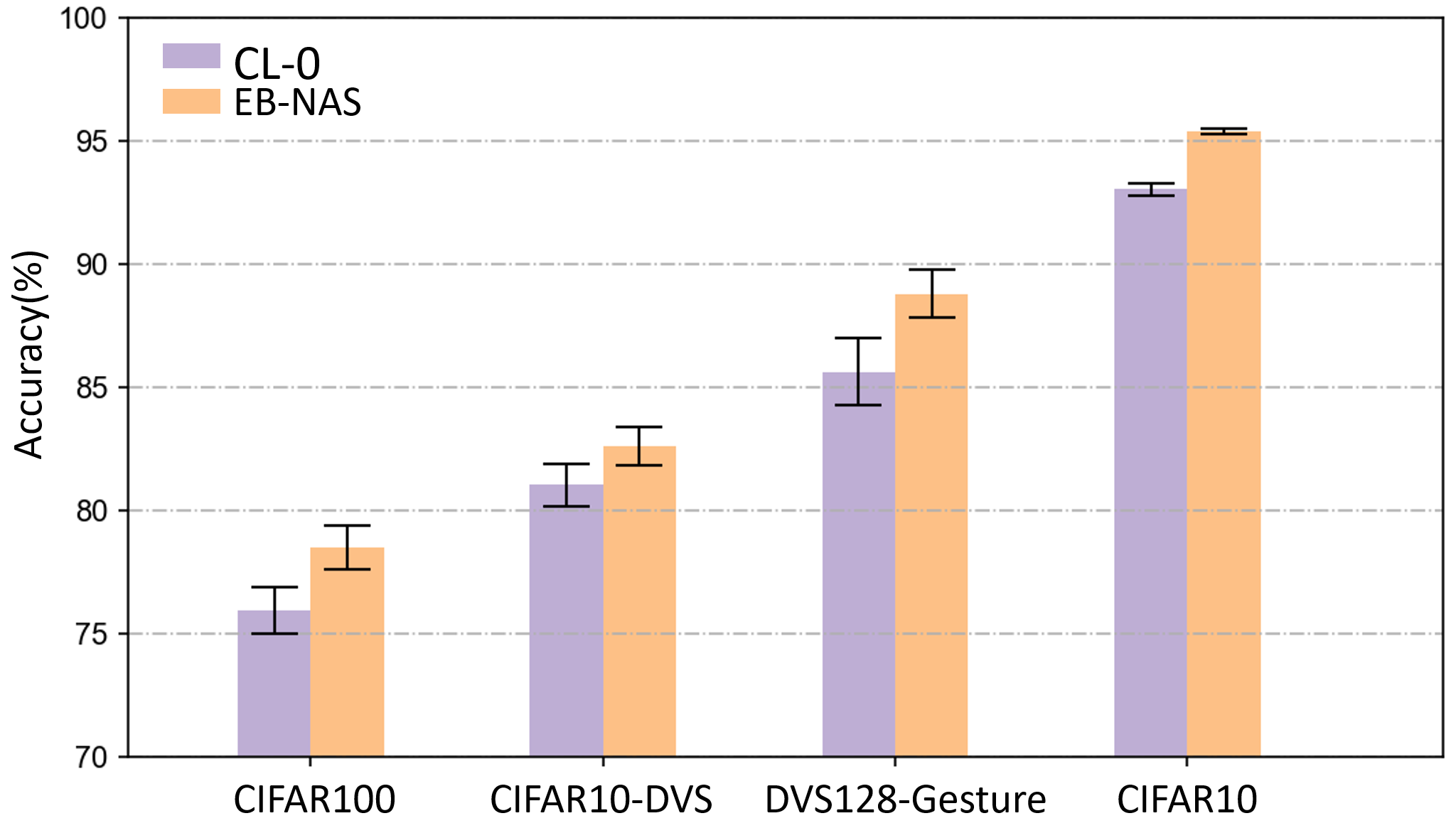}
\caption{Evolutionary performance comparison of CL-0 and EB-NAS. All candidates denoted by CL-0 are fixed to the hierarchical connection structure.}
\label{global}
\end{figure}

\subsubsection{Global Connections}
EB-NAS obtains free long-term connections between modules through evolution. 
We validate the role of this evolvable global connectivity on four datasets with two models: (1) the proposed method marked as EB-NAS; (2) the hierarchical connected model marked as CL-0. 
The ablation results are shown in Fig.~\ref{global}. It can be seen that the model evolved by the free global connection EB-NAS is 2.35\%, 2.54\%,  1.58\% and 3.17\% higher, respectively than the traditional forward connection CL-0 model on CIFAR10, CIFAR100, CIFAR10-DVS and DVS128-Gesture datasets.
The free global connection expands the search space of the traditional hierarchical-connected model, provides more possibilities and enriches the types of individuals that the evolutionary algorithm can search. This connectivity mode by imitating brain regions brings more features to each module and enhances the computing power of the whole model.

\subsection{Effects of Performance Predictor}


The performance predictor trains a regression model that can predict the accuracy of unknown genotypes through the knowledge of known genotypes and corresponding classification accuracy, which greatly shortens the time required for NSGA-\uppercase\expandafter{\romannumeral2}: in each iteration, all $n_{new}*gens=2400$ individuals originally needed to be trained for $e_{eval}$ epochs. 
With the help of the predictor, only $iters*k=500$ non-dominated models need to be trained, and the algorithm speed is 4.8x faster than the model without a predictor.

To demonstrate the performance of the predictor, we plot the correlation between the loss predicted by the predictor and the true loss during iteration on CIFAR10-DVS and DVS128-Gesture datasets, as Fig.~\ref{abl}b and c shown.
In Fig.~\ref{abl}b and c, the y-axis is the Spearman Correlation coefficient~\cite{spearman1961proof} between the prediction result of the predictor and the true performance of individuals. The larger the coefficient, the more accurate the prediction. 
It can be seen from the straight line fitting the scatter points that as the iteration progresses, the predictions of the predictor on both datasets are getting more and more accurate.

\subsection{Multi-objective Evolution Efficiency}
Hypervolume~\cite{zitzler2007hypervolume} represents the volume of the hypercube enclosed by the non-dominated solution set and the reference point in the target space~\cite{lin2021multi}, which can be used to evaluate the efficiency of multi-objective evolution. The higher the value, the more optimal solutions evolution brings to the population, i.e. a better Pareto front is obtained. 
We construct a baseline model for comparing evolution efficiency: random model. 
At this time, the $O$ to be input into $Set_1$ is randomly sampled in the entire search space, rather than the non-dominated solutions generated by NSGA-\uppercase\expandafter{\romannumeral2}.
The changing trend of hypervolume along with the evolution process on CIFAR10-DVS and DVS128-Gesture is drawn in Fig.~\ref{abl}d and e. 
The orange and blue lines represent the result of EB-NAS, and the grey line represents the result of the random model.
It can be seen that as the searching proceeds, the hypervolume of EB-NAS and random model are constantly increasing, and our method EB-NAS can reach larger hypervolume faster than the random model on both datasets.

In SNN, information is transmitted in the form of spikes, and fewer spikes mean more energy-efficient models.
By changing the initial channels of the model, we obtain different classification accuracy results evolved on CIFAR10, and compare it with other SNN models, as shown in Table~\ref{spikes}.
Compared with fixed-structure models, our model has obvious advantages in energy consumption. For example, we use no more than one-third of the number of spikes to achieve about 1\% higher performance than~\cite{fang2021incorporating}.
Compared with the same NAS-based method~\cite{na2022autosnn}, EB-NAS only consumes about half (178k) of the number of spikes (310k) to achieve an improvement of 2.39\%. Under the same energy consumption (178K and 176K), the performance of EB-NAS is 4.22\% higher.
\begin{table}[htp]
\centering
  \resizebox{3 in}{!}{
\begin{tabular}{llll}
\toprule
SNN Models &Accuracy (\%) & Spikes(K)\\
\midrule
\multirow{3}{*}{ CIFARNet-Wu~\cite{wu2019direct}} 
& 84.36 & $361$ \\
& 86.62 & $655 $ \\
& 87.80 & $1298 $ \\
\midrule

\multirow{5}{*}{ CIFARNet-Fang~\cite{fang2021incorporating}} 
 & 80.82 & $104 $ \\
 & 86.05 & $160 $ \\
& 90.83 & $260 $ \\
& 92.33 & $290 $ \\
& 93.15 & $507 $ \\
\midrule

\multirow{3}{*}{ ResNet11-Lee~\cite{lee2020enabling} } 
 & 84.43 & $140 $ \\
& 87.95 & $301 $ \\
& 90.24 & $ 1530 $ \\
\midrule

\multirow{4}{*}{ ResNet19-Zheng~\cite{zheng2021going}} 
 & 83.95 & $341$ \\
 & 89.51 & $541$ \\
& 90.95 & $853$ \\
& 93.07 & $1246$ \\
\midrule

\multirow{4}{*}{  AutoSNN~\cite{na2022autosnn}} 
& 88.67 & $108$ \\
 & 91.32 & $176$ \\
 & 92.54 & $261$ \\
 & 93.15 & $310$ \\

\midrule

\multirow{3}{*}{\textbf{ EB-NAS}} 
& \textbf{89.80}& $\bm{86}$ \\
 & \textbf{94.11} & $\bm{129} $\\
 &\textbf{95.54} &$\bm{178} $\\

\bottomrule

\end{tabular}}
\caption{Comparison result of spike numbers of EB-NAS with different accuracy on CIFAR10.}\label{spikes}
\end{table}

\section{Conclusion and future work}
The existing research on SNN is almost limited to the fixed structure inspired by DNN, and the performance of a few NAS-based SNN models does not exceed them, leaving much room for improvement. To address this challenge, we automatically evolve the architecture of SNNs from a brain-inspired perspective based on the combination of common neural motifs and free long-term connections between them. To accelerate the multi-objective evolution process, a time-saving online performance predictor is employed to find high-performance and low-energy SNN architectures.
Extensive ablation and comparison experiments show that the proposed biologically interpretable brain-inspired search space and online performance predictor can effectively improve the classification accuracy of candidate SNN architectures through neuroevolution and achieve state-of-the-art (SOTA) performance with significantly lower energy consumption, no matter compared with fixed-architecture SNNs or NAS-based SNNs. 
This not only demonstrates the powerful performance of EB-NAS but also reveals auto-evolving SNN architectures that can surpass human expert experience, unveiling the effectiveness of the natural evolution mechanism in the practice of SNN architecture design from the perspective of computational modelling.

As more and more mysteries of the human brain are uncovered, we expect to discover other brain-inspired topological properties that can guide the optimization of SNN's advanced cognitive functions, providing new clues for future exploration of other complex tasks (such as transfer learning, continuous learning and lifelong learning) and the realization of general artificial intelligence with adaptive multi-brain region evolution.

\section{Acknowledgement}
This work is supported by the National Key Research and Development Program (Grant No. 2020AAA0107800), the Strategic Priority Research Program of the Chinese Academy of Sciences (Grant No. XDB32070100),  the National Natural Science Foundation of China (Grant No. 62106261).

\bibliography{IEEEabrv,bibfile}

\section{Author Biography}
\begin{IEEEbiographynophoto}{Wenxuan Pan}received her B.Eng. degree from the University of Electronic Science and Technology of China, Chengdu, Sichuan, China. She is currently a Ph.D candidate in the Brain-inspired Cognitive Intelligence Lab, Institute of Automation, Chinese Academy of Sciences, supervised by Prof. Yi Zeng. Her research interests focus on brain-inspired evolutionary spiking neural networks.
\end{IEEEbiographynophoto} 
\begin{IEEEbiographynophoto}{Feifei Zhao}received the BS degree in digital media technology from Northeastern University, in 2014, and the PhD degree in pattern recognition and intelligent systems from the University of Chinese Academy of Sciences, in 2019. She is currently an Associate Professor with the Brain-inspired Cognitive Intelligence Lab, Institute of Automation, Chinese Academy of Sciences. Her research interests include multi-brain areas coordinated learning and decision-making spiking neural network, brain-inspired developmental and evolutionay architecture search for SNNs. \end{IEEEbiographynophoto}

\begin{IEEEbiographynophoto}{Zhuoya Zhao} received the B.Eng. degree from Harbin Engineering University in 2019.
Now she is the Ph.D. candidate in the Brain-inspired Cognitive Intelligence Lab, Institute of Automation, Chinese Academy of Sciences, supervised by Prof.Yi Zeng. 
Her current research interests are brain-inspired theory of mind spiking neural networks.
\end{IEEEbiographynophoto}
\begin{IEEEbiographynophoto}{Yi Zeng}  
Yi Zeng obtained his Bachelor degree in 2004 and Ph.D degree in 2010 from Beijing University of Technology, China. He is currently a Professor and Director in the Brain-inspired Cognitive Intelligence Lab, Institute of Automation, Chinese Academy of Sciences (CASIA), China. He is a Principal Investigator in the Center for Excellence in Brain Science and Intelligence Technology, Chinese Academy of Sciences, China, and a Professor in the School of Future Technology, and School of Humanities, University of Chinese Academy of Sciences, China. He is a Principle Investigator of the State Key Laboratory of Multimodal Artificial Intelligence Systems, Institute of Automation, Chinese Academy of Sciences, China. His research interests include brain-inspired Artificial Intelligence, brain-inspired cognitive robotics, ethics and governance of Artificial Intelligence, etc.

 \end{IEEEbiographynophoto}
\end{document}